\pdfoutput=1

\documentclass[11pt]{article}

\usepackage[hyperref]{emnlp2022}

\usepackage{times}
\usepackage{latexsym}

\usepackage[T1]{fontenc}

\usepackage[utf8]{inputenc}

\usepackage{microtype}
\usepackage{xspace}
\usepackage{graphicx}
\usepackage{amsmath, amssymb}
\usepackage{algorithm}
\usepackage{algorithmic}
\usepackage{multirow}
\usepackage{booktabs} 
\usepackage{makecell}
\usepackage{paralist} 
\newcommand{\ourmodel}{\textsc{Pro}\textsc{Gen}\xspace}

\title{\ourmodel: Progressive Zero-shot Dataset Generation \\ via In-context Feedback}

\author{
Jiacheng Ye$^{\spadesuit\diamondsuit}$\thanks{\, Work done while interning at Shanghai AI Lab.}, 
Jiahui Gao$^{\spadesuit}$, 
Jiangtao Feng$^{\diamondsuit}$,
Zhiyong Wu$^{\diamondsuit}$,\\
\textbf{Tao Yu}$^{\spadesuit\heartsuit}$,
\textbf{Lingpeng Kong}$^{\spadesuit\diamondsuit}$\\
$^\diamondsuit$Shanghai AI Laboratory \quad
$^\heartsuit$University of Washington \\
$^\spadesuit$The University of Hong Kong \\
\texttt{\{carsonye, sumiler\}@connect.hku.hk,}  \\
\texttt{\{fengjiangtao,wuzhiyong\}@pjlab.org.cn, \{tyu,lpk\}@cs.hku.hk}
}

\begin{document}
\maketitle

\begin{abstract}
 Recently, dataset-generation-based zero-shot learning has shown promising results by training a task-specific model with a dataset synthesized from large pre-trained language models (PLMs). 
 The final task-specific model often achieves compatible or even better performance than PLMs under the zero-shot setting, with orders of magnitude fewer parameters.
However, synthetic datasets have their drawbacks. They have long been suffering from low-quality issues (e.g., low informativeness and redundancy). This explains why the massive synthetic data does not lead to better performance -- a scenario we would expect in the human-labeled data. 
To improve the quality of dataset synthesis, we propose a progressive zero-shot dataset generation framework, \ourmodel, which leverages the feedback from the task-specific model to guide the generation of new training data via in-context examples.
Extensive experiments on five text classification datasets demonstrate the effectiveness of the proposed approach. 
We also show \ourmodel achieves on-par or superior performance with only 1\% synthetic dataset size compared to baseline methods without in-context feedback.

\end{abstract}
\section{Introduction}
Dataset generation with pre-trained language models (PLMs) has attracted enormous interest recently due to the superior generative capacity of PLMs. Given task-specific supervision, recent work
\citep[\emph{inter alia}]{DBLP:conf/aaai/Anaby-TavorCGKK20,DBLP:conf/emnlp/PuriSSPC20,DBLP:journals/corr/abs-2003-02245,DBLP:journals/corr/abs-2102-01335} manages to fine-tune the PLMs to synthesize high-quality datasets for downstream applications. Nevertheless, obtaining task supervision from human experts can be expensive or even unrealistic. 
Recent attempts~\citep[\emph{inter alia}]{DBLP:conf/emnlp/SchickS21a,DBLP:journals/corr/abs-2109-09193,meng2022generating} turn their eyes to the unsupervised dataset generation. Among them, \textsc{ZeroGen} \citep{DBLP:journals/corr/abs-2202-07922} proposes to first convert the task descriptions into carefully designed prompts~\citep{DBLP:conf/emnlp/PetroniRRLBWM19,DBLP:conf/nips/BrownMRSKDNSSAA20}, and then use these prompts to steer the PLMs to synthesize the training data for the final task model. This approach allows highly efficient inference as the final task model only has orders of magnitude fewer parameters compared to PLMs, yet achieves compatible or even better performance than PLMs under the zero-shot setting. 
\begin{figure}[t]
\centering
\includegraphics[width=3in]{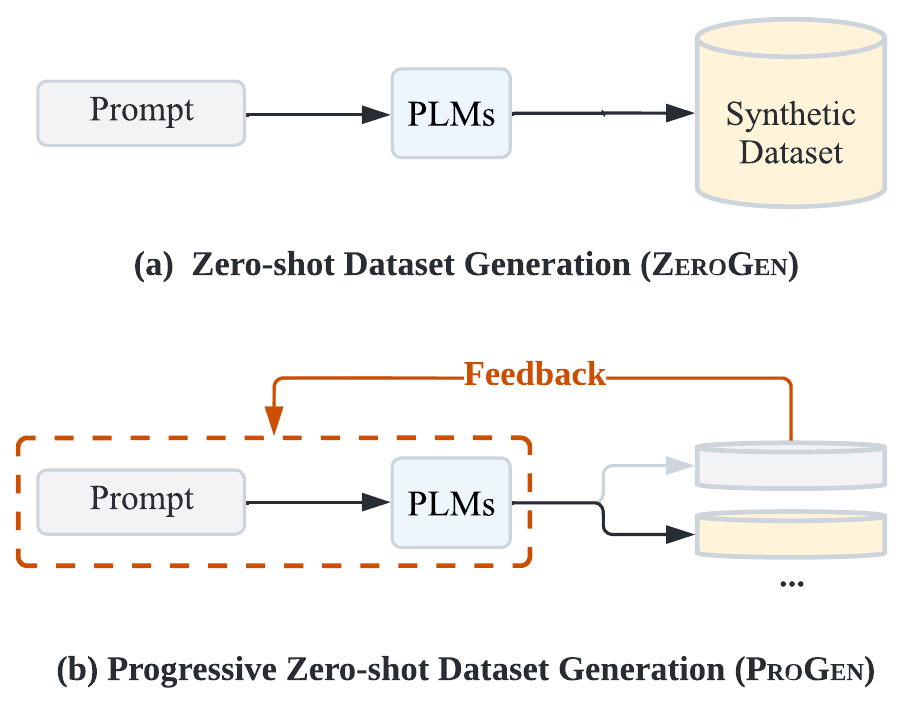}
\caption{Comparison of vanilla zero-shot dataset generation (\textsc{ZeroGen}) and progressive zero-shot dataset generation  (\textsc{ProGen}). In progressive zero-shot dataset generation, we split the whole dataset generation process into multiple phrases. In each phase, the generation is steered by feedback from the previously generated dataset, so as to synthesize a dataset with higher quality.
}
\label{fig:idg}
\end{figure}

The major drawback of synthetic datasets, however, is they often suffer from low-quality issues (e.g., low informativeness, redundancy). Despite we can generate as much data as computational resource allows, the massive generated data does not automatically translate into better performances, unlike in the human-labeling scenario.

To address this problem, we propose a \underline{pro}gressive zero-shot dataset generation framework (Figure~\ref{fig:idg}b), called \ourmodel.
In a nutshell, \ourmodel learns a model for a downstream task by performing two phrases alternatively -- using PLMs to create labeled examples leveraging the feedback from the current task-specific model, and training a task-specific model given the generated labeled examples.
To compute reliable signals as feedback, we employ the influence function (\citet{DBLP:conf/icml/KohL17}; IF)
to quantify contribution to the loss for each training point. 
In the context of zero-shot learning where no human-annotated data is assumed,
we integrate a noise-resistant objective in the calculation of IF so that it can tackle the noise in the synthetic dataset.
To incorporate feedback into PLMs, we sort the training samples based on their quantified influence score, and formulate those most influential ones as in-context examples~\citep{DBLP:conf/nips/BrownMRSKDNSSAA20} to steer the generation. 
Overall, \ourmodel has the following advantages: 1) the quality estimation phrase requires no human annotations, thus works in a purely zero-shot learning setting; 2) unlike most  controllable generation methods that tune or require the access to PLMs~\citep[\emph{inter alia}]{DBLP:journals/corr/abs-1909-05858,DBLP:conf/iclr/DathathriMLHFMY20,DBLP:conf/acl/LiuSLSBSC20}, the in-context feedback phrase does not need to modify parameters in the PLM and incurs minimal disturbance to its generation procedure.
Our main contributions are three folds:
\begin{itemize}
\item We propose a progressive framework for zero-shot dataset generation to generate higher-quality dataset (\S\ref{sec:progen});
\item We propose noise-resistant influence function to estimate the quality of each sample without any human annotations (\S\ref{sec:if_quality}), and a learning-free controllable generation method via in-context feedback (\S\ref{sec:feedback});
\item Across multiple text classification datasets, we show our framework obtains better performance over various prompt-based methods, and achieves on-par zero-shot performance with only 1\% synthetic dataset size,
when compared to methods without in-context feedback (\S\ref{sec:exp}).
\end{itemize}
Our code can be found at \url{https://github.com/HKUNLP/ProGen}.

\begin{figure*}[t]
\centering
\includegraphics[width=6in]{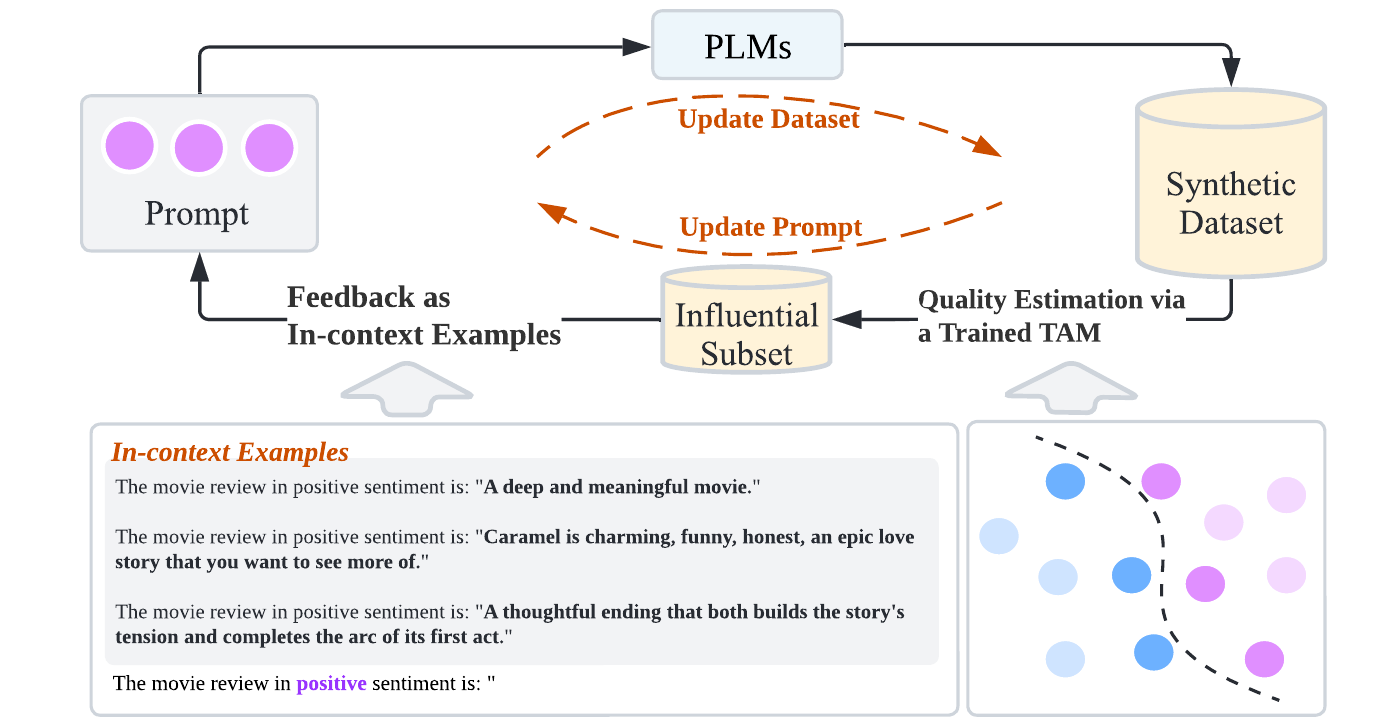}
\caption{Framework of \ourmodel for progressive zero-shot dataset generation. 
To update the prompt, we first train a task-specific model (TAM) with the synthetic dataset, and then employ the noise-robust influence function to measure the quality of each data point. Finally, the most influential subset is selected, which acts as feedback via in-context learning.
The whole framework works with a black-box PLM and requires no human annotations.
}
\label{fig:model}
\end{figure*}

\section{Background}
\label{sec:zerogen}
In this section, we briefly review the baseline approaches of zero-shot dataset generation and how the synthesized dataset can be used for zero-shot learning on downstream tasks. 


\paragraph{Zero-shot Dataset Generation}
Take text classification task as an example, vanilla zero-shot dataset generation methods~\citep{meng2022generating,DBLP:journals/corr/abs-2202-07922} aims to generate a synthetic dataset $\mathcal{D} = \{(\mathbf{x}, y)\} $ with the help of a PLM $\mathcal{P}$. 
They first sample a class label $y$ from a uniform distribution: 
\begin{equation}
y \sim \mathbf{U}(y_1, y_2, \ldots, y_k),
\end{equation}
where $k$ is the number of classes. They then wrap $y$ up into a label-descriptive prompt $\mathcal{T}(y)$ to steer the generation of $\mathbf{x}$: 
\begin{equation}
\mathbf{x} \sim \mathcal{P}(\cdot|\mathcal{T}(y)).
\end{equation}
Since the parameters of $\mathcal{P}$ is frozen and the generation $\mathbf{x}$ for each $y$ is deterministic, different sampling algorithms (e.g., Top-k sampling \cite{DBLP:conf/acl/LewisDF18} and nucleus sampling \cite{DBLP:conf/iclr/HoltzmanBDFC20}) can be adopted to increase the diversity of generated dataset. A synthetic dataset $\mathcal{D}$ is constructed after pairing the generated $\mathbf{x}$ with $y$.  

\paragraph{Dataset-generation-based Zero-shot Learning}
The vast linguistic~\citep{DBLP:conf/acl/JawaharSS19,DBLP:journals/corr/abs-1901-05287,DBLP:conf/iclr/TenneyXCWPMKDBD19} and factual~\citep{DBLP:conf/emnlp/PetroniRRLBWM19,DBLP:journals/tacl/JiangXAN20} knowledge encoded in PLMs' parameters is the key towards the success of conventional prompt-based zero-shot learning (\textsc{Prompting})~\citep{DBLP:conf/nips/BrownMRSKDNSSAA20}. However, \textsc{Prompting} fails to fully exert the capacity of PLMs and heavily relies on gigantic PLMs during inference. This motivates another line of work~\citep{meng2022generating,DBLP:journals/corr/abs-2202-07922} to explore a more flexible and efficient way of conducting zero-shot learning based on dataset generation. Given the synthetic dataset generated as above, a task-specific model is trained, allowing any task-specific inductive bias and with an order-of-magnitude smaller number of parameters compared to PLMs. 
The performance of the final task-specific model is mostly dominated by the quality of the synthetic dataset, and a low-quality dataset degrades the final zero-shot performance. This thereby motivates us to explore methods that improve the dataset quality.


\section{\ourmodel}
\label{sec:progen}
We now describe our framework for \underline{pro}gressive zero-shot dataset \underline{gen}eration via in-context feedback (\ourmodel), as shown in Figure~\ref{fig:model}. We follow \textsc{ZeroGen}~\citep{DBLP:journals/corr/abs-2202-07922} to build the backbone of our framework. Concretely, we first train a \underline{ta}sk-specific \underline{m}odel (TAM) with partially generated dataset. Then, assuming no access to human annotations, we estimate the influence of each sample via the noise-robust influence function. Finally, with those identified most influential samples, we explore the use of in-context learning to shift the generation distribution towards that of influential samples, so that the system generates more related samples. The whole framework progressively constructs the synthetic dataset and enhances the performance of the final task-specific model.

\subsection{Annotation-free Quality Estimation}
\label{sec:if_quality}
There are many factors in measuring the quality of a dataset, e.g., diversity, annotation correctness, spurious biases~\citep{DBLP:journals/corr/abs-2005-00816,DBLP:journals/corr/abs-2102-12060}.
However, it is often very subjective, making it unrealistic to calculate them all automatically. 
Our solution to this is to infer the quality of the individual samples in synthetic datasets using the performance of the final task-specific model trained on the dataset as the surrogate.
Concretely, we propose to apply influence function~\citep{DBLP:conf/icml/KohL17} on the task-specific model to give sample-level influence scores with regard to the loss of validation set. However, a clean validation set, which is crucial for producing reliable influence scores, is inaccessible in the zero-shot learning setting. Thus, we use a synthetic validation set and harness the influence function with a noise-robust objective to handle the potential noise in the synthetic validation set.

Formally, influence function measures the change in the model's loss on the test data-point $z_{\text{test}}=(\mathbf{x}, y)$ if we up-weight the loss of a training data-point $z$ by $\epsilon$:
\begin{equation}
\small
\begin{aligned}
\mathcal{I}_{\text {up,loss }}\left(z, z_{\text {test }}\right) &\left.\stackrel{\text {def}}{=} \frac{d L\left(z_{\text {test}}, \hat{\theta}_{\epsilon, z}\right)}{d \epsilon}\right|_{\epsilon=0} \\
&=\left.\nabla_{\theta} L\left(z_{\text {test}}, \hat{\theta}\right)^{\top} \frac{d \hat{\theta}_{\epsilon, z}}{d \epsilon}\right|_{\epsilon=0} \\
&=-\nabla_{\theta} L\left(z_{\text {test}}, \hat{\theta}\right)^{\top} H_{\hat{\theta}}^{-1} \nabla_{\theta} L(z, \hat{\theta}),
\end{aligned}
\end{equation}
where $\hat{\theta}_{\epsilon, z} \stackrel{\text {def}}{=} \arg \min _{\theta \in \Theta} \frac{1}{n} \sum_{i=1}^{n} L\left(z_{i}, \theta\right)+\epsilon L(z, \theta)$ is the parameter if $z$ were upweighted by some small $\epsilon$ and $H_{\hat{\theta}}$ is the Hessian. Our noise-robust validation-set level influence function is defined as:
\begin{equation}
\label{eq:if}
\small
\mathcal{I}_{\text {up,loss}}\left(z, \mathcal{D}_{\text {val}}\right) = -\nabla_{\theta} L^\prime \left(\mathcal{D}_{\text {val}}, \hat{\theta}\right)^{\top} H_{\hat{\theta}}^{-1} \nabla_{\theta} L(z, \hat{\theta}),
\end{equation}
where $L^\prime$ is a noise-tolerant loss. In this work, we adopt Reverse Cross-Entropy (RCE) loss~\citep{wang2019symmetric}, which has the following form:
\begin{equation}
\small
L^\prime \left(\mathcal{D}_{\text {val}}, \hat{\theta}\right)=\sum_{i=1}^{|\mathcal{D}_{\text{val}}|} \ell_{\hat{\theta}}(\hat{y}_i, y_i)=-\sum_{i=1}^{|\mathcal{D}_{\text{val}}|} \sum_{c=1}^C\hat{y}_i^c\log(y_i^c),     
\end{equation}
where $\hat{y}_i$ is the predicted class for sample $i$, $C$ is the number of classes.\footnote{$\log(0)$ is approximated to a constant $A$ in the case of $y_i^c=0$.} A smaller negative value of $\mathcal{I}_{\text {up,loss}}$ indicates that upweighting the corresponding training sample will decrease the validation loss more, thus the sample is more valuable. After sorting training samples with $\mathcal{I}_{\text {up,loss}}$ ascendingly, we select top-$M$ (e.g., 50) most valuable samples to form a tiny dataset, which we 
denote as $\mathcal{D}_{\text{helpful}}=\{(\mathbf{x}, y)\}$. 

In practice, we randomly sample a subset of $\mathcal{D}_{\text{train}}$ and adopt the stochastic estimation method described in~\citet{DBLP:conf/icml/KohL17} to efficiently compute $\mathcal{I}_{\text {up,loss}}$. 

\subsection{Feedback via In-context Learning}
\label{sec:feedback}
After identifying influence scores for each previously generated sample, we hypothesize that including additional samples similar to those most helpful ones can boost downstream performance. 
Instead of purely paraphrasing those helpful samples individually, which may hinder the diversity of synthetic dataset, we expect the model to learn from the overall distribution of those helpful samples and generate new samples of similar quality. 

Motivated by the striking capability of the in-context Learning~\citep{DBLP:conf/nips/BrownMRSKDNSSAA20} for PLMs, we propose
to use the identified important samples as in-context examples, so that they can shift the generation distribution of PLMs to the ones that are more beneficial to the training of the final task-specific model. Formally, each $\mathbf{x}$ is now generated as follows: 
\begin{equation}
\label{eq:in-context}
\mathbf{x} \sim \mathcal{P}(\cdot|\mathcal{T}( y_1, \mathbf{x}_1),\dots,\mathcal{T}(y_k,\mathbf{x}_k),\mathcal{T}(y)),
\end{equation}
where all the in-context examples $\{(\mathbf{x}_i,y_i)\}_{i=1}^k$ are randomly selected from $\mathcal{D}_{\text{helpful}}$. Compared with controllable text generation methods that directly modify the parameters in the PLM, we argue that the in-context learning methods incur minimal disturbance to the model's generation procedure.

The overall framework of \ourmodel is elaborated in Algorithm~\ref{alg}. 

\begin{algorithm}[htb!]
\caption{Progressive Zero-shot Dataset Generation}\label{alg}
\begin{algorithmic}[1]
\REQUIRE a PLM, a TAM, feedback interval $I$, iterations $T$.
\STATE $\mathcal{D}_{\text{train}}$ $\leftarrow$ $\varnothing$
\STATE $\mathcal{D}_{\text{helpful}}$ $\leftarrow$ $\varnothing$
\STATE $\mathcal{D}_{\text{val}}$ $\leftarrow$ Generate a validation set with PLM.
\FOR{feedback iteration $t = 1, 2 \ldots T$}
\STATE $\mathcal{D}_{\text{new}}$ $\leftarrow$ Generate a dataset of size $I$ with PLM and $\mathcal{D}_{\text{helpful}}$ via Eqn.~\ref{eq:in-context}.
\STATE $\mathcal{D}_{\text{train}}$ $\leftarrow$ $\mathcal{D}_{\text{train}}$ $\cup$ $\mathcal{D}_{\text{new}}$.
\STATE TAM $\leftarrow$ Training with  $\mathcal{D}_{\text{train}}$ and $\mathcal{D}_{\text{val}}$.
\STATE $\mathcal{D}_{\text{helpful}}$ $\leftarrow$ Select most helpful subset from $\mathcal{D}_{\text{train}}$ with TAM and $\mathcal{D}_{\text{val}}$ via Eqn.~\ref{eq:if}.
\ENDFOR

\textbf{Output:} $\mathcal{D}_{\text{train}}$.
\end{algorithmic}
\end{algorithm}
\section{Experiments}
\label{sec:exp}
\subsection{Setup}

\paragraph{Datasets}
We evaluate our method on five natural language text classification datasets, including IMDb~\citep{DBLP:conf/acl/MaasDPHNP11}, SST-2~\citep{DBLP:conf/emnlp/SocherPWCMNP13}, Rotten Tomatoes~\citep{pang-lee-2005-seeing}, Elec~\citep{McAuley2013} and Yelp~\citep{zhang_yelp}. Among these datasets, 
IMDB, SST-2, and Rotten Tomatoes are binary classification benchmarks for movie reviews, Elec and Yelp are binary classification tasks for electronic product reviews and restaurant reviews, respectively. 
The sizes of the training and test set are 25k/25k, 	
6.9k/0.8k, 8.5k/1k, 25k/25k, and 560k/38k for IMDb, SST-2, Rotten Tomato, Elec and Yelp, respectively.

\begin{table*}[t]
\centering
\small
\scalebox{0.92}{
\begin{tabular}{lll|cccccc}
\toprule
 \textbf{TAM} & {\textbf{\#Param}} & {\textbf{Setting}} &\textbf{IMDb} & \textbf{SST-2} & \textbf{Rotten Tomato} & \textbf{Elec} & \textbf{ Yelp } & \textbf{ Avg. }\\
 \hline
 \#\textit{Gold Data} & &\multirow{3}{*}{{\textsc{Supervised}}} & 25k & 6.7k & 8.3k & 25k & 560k & -\\
 DistilBERT & 66M & & 87.24\phantom{\scriptsize $\pm 0.00$} & 89.68\phantom{\scriptsize $\pm 0.00$} & 83.67\phantom{\scriptsize $\pm 0.00$} & 92.63\phantom{\scriptsize $\pm 0.00$} & 95.42\phantom{\scriptsize $\pm 0.00$} &89.73\phantom{\scriptsize $\pm 0.00$}\\
LSTM & $\sim$7M &   & 84.60\phantom{\scriptsize $\pm 0.00$} & 76.30\phantom{\scriptsize $\pm 0.00$} & 77.49\phantom{\scriptsize $\pm 0.00$} & 86.36\phantom{\scriptsize $\pm 0.00$} &91.30\phantom{\scriptsize $\pm 0.00$} &83.21\phantom{\scriptsize $\pm 0.00$}  \\
\hline
\hline
\multirow{2}{*}{-} & \multirow{2}{*}{1.5B} &  {{\textsc{Prompting}}} &  
70.50\textcolor{gray}{\scriptsize $\pm 14.3$} & 71.05\textcolor{gray}{\scriptsize $\pm 26.0$} & 68.58\textcolor{gray}{\scriptsize $\pm 22.2$} & 72.76\textcolor{gray}{\scriptsize $\pm 6.62$} & 75.52\textcolor{gray}{\scriptsize $\pm 10.2$} & 71.68\textcolor{gray}{\scriptsize $\pm 15.9$} \\
& & {{\textsc{Prompting}$^*$}} & 
77.31\textcolor{gray}{\scriptsize $\pm 2.23$} & 82.63\textcolor{gray}{\scriptsize $\pm 8.35$} & 78.66\textcolor{gray}{\scriptsize $\pm 7.23$} & 78.03\textcolor{gray}{\scriptsize $\pm 2.29$} & 80.30\textcolor{gray}{\scriptsize $\pm 6.69$} & 79.39\textcolor{gray}{\scriptsize $\pm 5.36$} \\
\hline
 \multirow{2}{*}{DistilBERT} & \multirow{2}{*}{66M} &  \textit{\textsc{ZeroGen}} &   80.41\textcolor{gray}{\scriptsize $\pm 5.38$} & 82.77\textcolor{gray}{\scriptsize $\pm 6.24$} & 78.36\textcolor{gray}{\scriptsize $\pm 7.68$} & 85.35\textcolor{gray}{\scriptsize $\pm 3.07$} & 87.84\textcolor{gray}{\scriptsize $\pm 2.45$} & 82.94\textcolor{gray}{\scriptsize $\pm 4.96$} \\
 &  &\textit{\ourmodel} & \bf{84.12\textcolor{gray}{\scriptsize $\pm 0.26$}}  & \bf{87.20\textcolor{gray}{\scriptsize $\pm 1.21$}} & \bf{82.86\textcolor{gray}{\scriptsize $\pm 1.27$}} & \bf{89.00\textcolor{gray}{\scriptsize $\pm 1.16$}} & \bf{89.39\textcolor{gray}{\scriptsize $\pm 0.30$}} & \bf{86.51\textcolor{gray}{\scriptsize $\pm 0.84$}} \\
 \hline
 \multirow{2}{*}{LSTM} & \multirow{2}{*}{$\sim$7M} &  \textit{\textsc{ZeroGen}} & 70.18\textcolor{gray}{\scriptsize $\pm 8.53$} & 75.53\textcolor{gray}{\scriptsize $\pm 10.1$} & 72.48\textcolor{gray}{\scriptsize $\pm 9.36$} & 75.84\textcolor{gray}{\scriptsize $\pm 5.74$} & 83.75\textcolor{gray}{\scriptsize $\pm 2.17$} & 75.56\textcolor{gray}{\scriptsize $\pm 7.19$} \\
 & & \textit{\ourmodel} & 
 \bf{77.85\textcolor{gray}{\scriptsize $\pm 0.84$}} & 
 \bf{80.96\textcolor{gray}{\scriptsize $\pm 1.78$}} & 
 \bf{77.27\textcolor{gray}{\scriptsize $\pm 1.51$}} & 
 \bf{82.85\textcolor{gray}{\scriptsize $\pm 3.17$}} & \bf{86.03\textcolor{gray}{\scriptsize $\pm 1.62$}} & 
 \bf{80.99\textcolor{gray}{\scriptsize $\pm 1.78$}}
 \\
\bottomrule
\end{tabular}}
\caption{Evaluation results with two different scales of TAM. The scale of synthetic dataset is 100k for both \textsc{ZeroGen} and \ourmodel. We report the average accuracy and corresponding standard deviation across multiple prompts. The detailed results for each prompt are shown in Appendix~\ref{app:full_results}.}
\label{tab:main}
\end{table*}

\paragraph{Evaluation Strategy}
Following~\citet{DBLP:journals/corr/abs-2202-07922}, we evaluate the quality of the synthetic dataset by first training a task-specific model (TAM) with the dataset, and then testing it on a human-annotated dataset (i.e., test set). We also explore other evaluation metrics in \S~\ref{sec:analysis}.

\paragraph{Baselines}
The TAM trained with the synthetic dataset can perform zero-shot inference, hence, we compare \ourmodel with other zero-shot learning baselines:
\begin{itemize}
    \item \textsc{Prompting}. The prompt-based zero-shot classification method based via PLMs~\citep{DBLP:conf/nips/BrownMRSKDNSSAA20}. 
    \item \textsc{Prompting}$^*$. The calibrated prompting method that reweighs each option according to its priori likelihood~\citep{DBLP:conf/emnlp/HoltzmanWSCZ21}. 
    \item \textsc{ZeroGen}. A recent zero-shot learning work via dataset generation~\citep{DBLP:journals/corr/abs-2202-07922}.
    They first generate a dataset with a carefully designed instruction, and then train a tiny task-specific model (TAM) to conduct zero-shot inference.
\end{itemize}
We also provide a non-zero-shot learning baseline \textsc{Supervised} where the same TAM is used but trained on the human-annotated training set.

\paragraph{Implementation Details}
Following~\citet{DBLP:journals/corr/abs-2202-07922},
we use GPT2-XL~\citep{radford2018improving} and Nucleus Sampling~\citep{DBLP:conf/iclr/HoltzmanBDFC20} with $p=0.9$ for dataset generation. 
Regarding prompt selection, we adopt a series of prompts for each task. The details of prompt selection are provided in Appendix~\ref{app:prompt}. 
By default, the feedback interval $I$ is set to 1k, and iteration $T$ is set to 100, which ends up with a dataset of size 100k in total. Calculating IF score for all training points is computationally expensive, thus we only sample 10k samples in each iteration. 
In practice, we find generate data using feedback all the time hinders diversity, thus we only apply feedback half of the time. 

We implement an LSTM-based model and a DistilBERT model as TAM to measure the quality of the synthetic dataset. For the LSTM-based model, we use Adam optimizer~\citep{DBLP:journals/corr/KingmaB14}, a learning rate of 1e-3, an embedding dim of 100, a hidden size of 300, and a layer number of 1.
For DistilBERT, we fine-tune on each dataset with Adam optimizer, with a learning rate of 2e-5, a weight decay of 0.01, and other default hyper-parameters as suggested by HuggingFace Transformers library~\citep{DBLP:journals/corr/abs-1910-03771}.
While using stochastic estimation in the influence function, we randomly sample 10k samples from the whole synthetic dataset, and calculate influence score for those samples over the whole validation set. This operation roughly costs 7 minutes.
For all the experiments, we run on a single NVIDIA A100 GPU, and generating 100k examples cost 28h on average for \ourmodel.

\subsection{Main Results}
We evaluate the generated datasets by training two different task-specific models and testing their performance on multiple downstream tasks. The results are shown in Table~\ref{tab:main}. 
We find that \textsc{Prompting} suffers from high variance over various prompts in zero-shot learning, and \textsc{Prompting}$^*$ substantially improves average accuracy and reduces variance across different choices of the prompt through calibration, which is also observed by previous work~\citep{DBLP:conf/icml/ZhaoWFK021,DBLP:conf/emnlp/HoltzmanWSCZ21,DBLP:conf/acl/MinLHZ22}. 
Compared with \textsc{Prompting} and \textsc{Prompting}$^*$, \textsc{ZeroGen} achieves superior performance by distilling task-related knowledge through dataset generation and handling the downstream tasks with a discriminator rather than a generator. 
Despite \textsc{ZeroGen}'s success,  \ourmodel further boosts both average accuracy and variance by improving the quality of the synthesized dataset via a generate-then-feedback framework.

\subsection{Ablations}
\begin{table}[t]
\centering
\scalebox{0.85}{
\begin{tabular}{l|ccc}
\toprule
 & \multicolumn{1}{l}{\textbf{SST-2}} & \multicolumn{1}{l}{\textbf{Elec}} & \multicolumn{1}{l}{\textbf{Yelp}} \\
 \hline
Baseline (\textsc{ZeroGen}) & 82.77 & 85.35 & 87.84 \\
$+$ syn-random & 86.81 & 87.94 & 87.96 \\
$+$ syn-helpful$_{\text{ce}}$ & 86.77 & 87.74 & 89.12 \\
$+$ syn-helpful$_{\text{rce}}$(\ourmodel) & \textbf{87.20} & \textbf{89.00} & \textbf{89.39} \\
\hline
$+$ gold & 90.20 & 91.02 & 91.43 \\
\bottomrule
\end{tabular}}
\caption{Evaluation results when harnessing baseline with different types of in-context examples. \textbf{syn-random}: random generated samples. \textbf{syn-helpful$_{\text{ce}}$}: generated samples selected by influence function with cross-entropy loss. \textbf{syn-helpful$_{\text{rce}}$}: generated samples selected by influence function with reverse cross-entropy loss. \textbf{gold}: test set examples. We report the average results across multiple prompts.}
\label{tab:ablation}
\end{table}
One of the main contributions of the proposed progressive zero-shot dataset generation framework is that it incorporates the previously generated dataset as feedback to steer generation. 
We provide an ablation study over various types of feedback, and summarize the results in Table~\ref{tab:ablation}. 
We find that providing random synthetic examples as feedback consistently outperforms the baseline method. Our hypothesis is that  task-related in-context examples may demonstrate further task-related information than the prompt, and thus benefit the generation process. 
Selecting examples with vanilla influence function (i.e., CE), only achieves on-par performance with random selection, due to the fact that the signal comes from the noisy validation set is not reliable.\footnote{It assumes the validation set objective is the same as the training set.}
In contrast, applying a noise-tolerant objective (i.e., RCE) on the validation set achieves superior performance, which is resistant to the noise in validation set and is able to find more accurate important examples. This proves better task-related signals can further improve the generation quality. 
Moreover, we find selecting in-context examples from test set examples.\footnote{To prevent trivial solutions (i.e., directly copying in-context examples), we remove the generated texts that are highly overlapped with any given in-context examples.} obtains the best results, which indicates the model does learn from better in-context examples.

\subsection{Analysis}
\label{sec:analysis}

\paragraph{Noise-tolerant influence function provides better estimation in a noisy-validation set scenario.}
\label{sec:if}
\begin{figure}[t]
\centering
\includegraphics[width=3.15in]{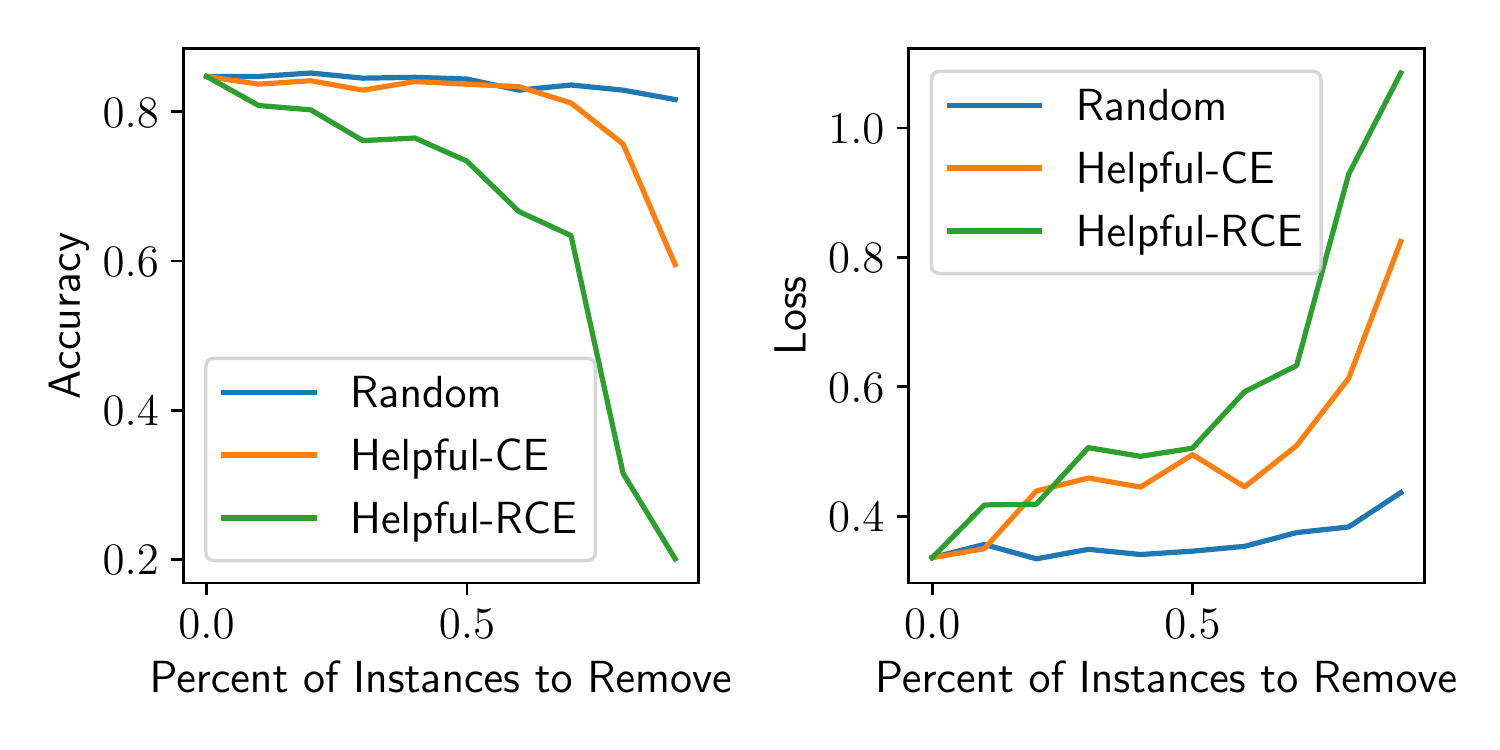}
\caption{Comparison of various ways to select important examples and the corresponding effects on test set accuracy and loss when removing them. \textbf{Helpful-CE}: helpful examples identified by vanilla influence function with cross-entropy loss. \textbf{Helpful-RCE}: helpful examples identified by robust influence function with reverse cross-entropy loss.
}
\label{fig:if}
\end{figure}

To see whether using a robust loss function on the validation set contributes to a more accurate estimate, we use a fixed synthetic dataset, remove the estimated important examples, and show how the accuracy and loss of a task-specific model trained with the remained dataset changes. We study three estimation methods, various remove ratios, and evaluate on SST-2 gold test set, as shown in Figure~\ref{fig:if}. We find filtering random synthetic examples almost does not hurt accuracy and achieves similar accuracy with only 10\% examples. Removing helpful examples identified by influence function with cross-entropy increases loss to some extent, but the degree of change is less than using reverse cross-entropy. This shows reverse cross-entropy augmented influence function could offer a more accurate estimate. We also compare results on the test set and artificial noisy test set as validation set in Appendix~\ref{app:artificial}, and demonstrate that the two losses are similar when the validation set is clean, but reverse cross-entropy is more effective when the validation set is noisy. 

\paragraph{Format of in-context examples is important. }
\label{sec:format}

\begin{figure}[t]
\centering
\includegraphics[width=3.15in]{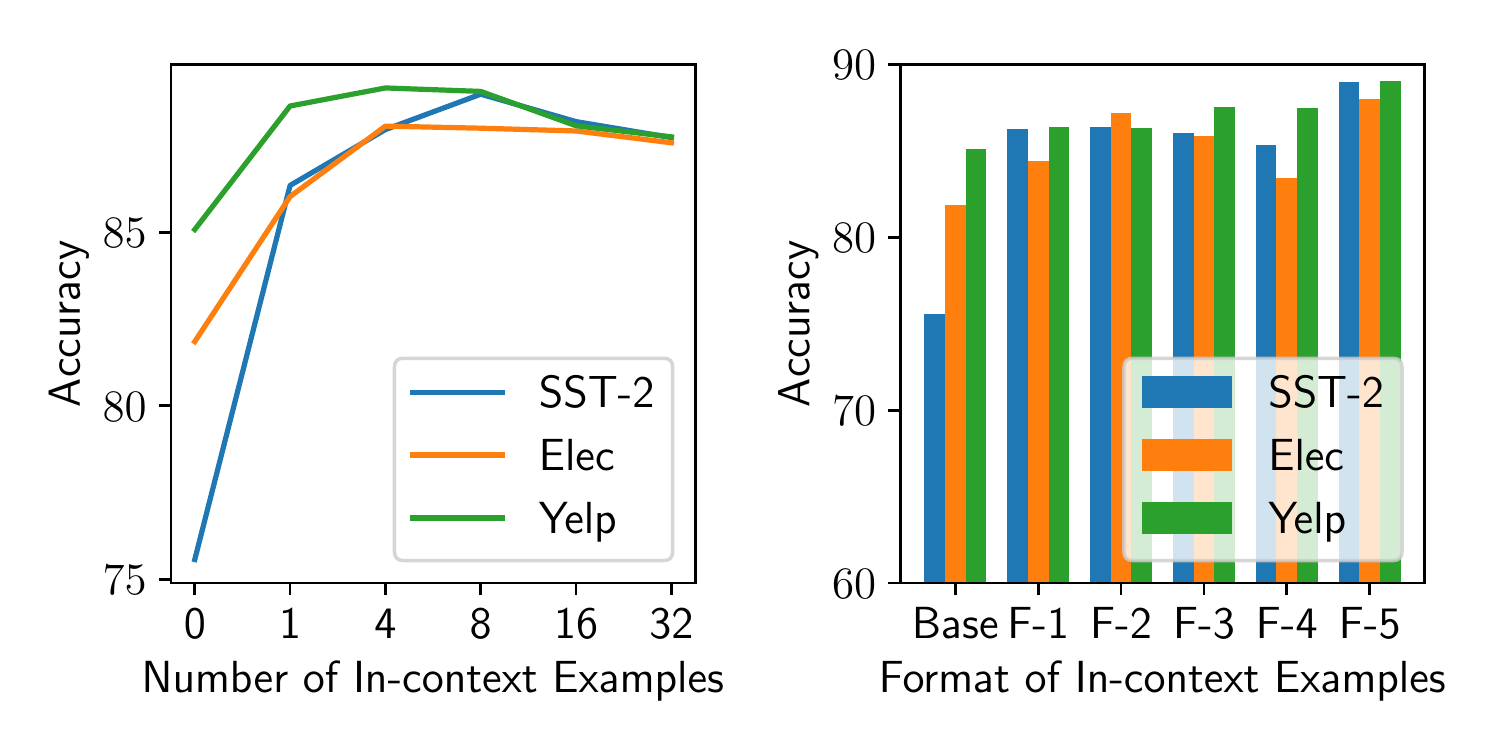}
\caption{Comparison of different number and format of in-context examples. F-$*$ are different format of in-context examples, see \S~\ref{sec:format} in detail.
}
\label{fig:format}
\end{figure}

\begin{figure*}[ht]
\centering
\includegraphics[width=5.5in]{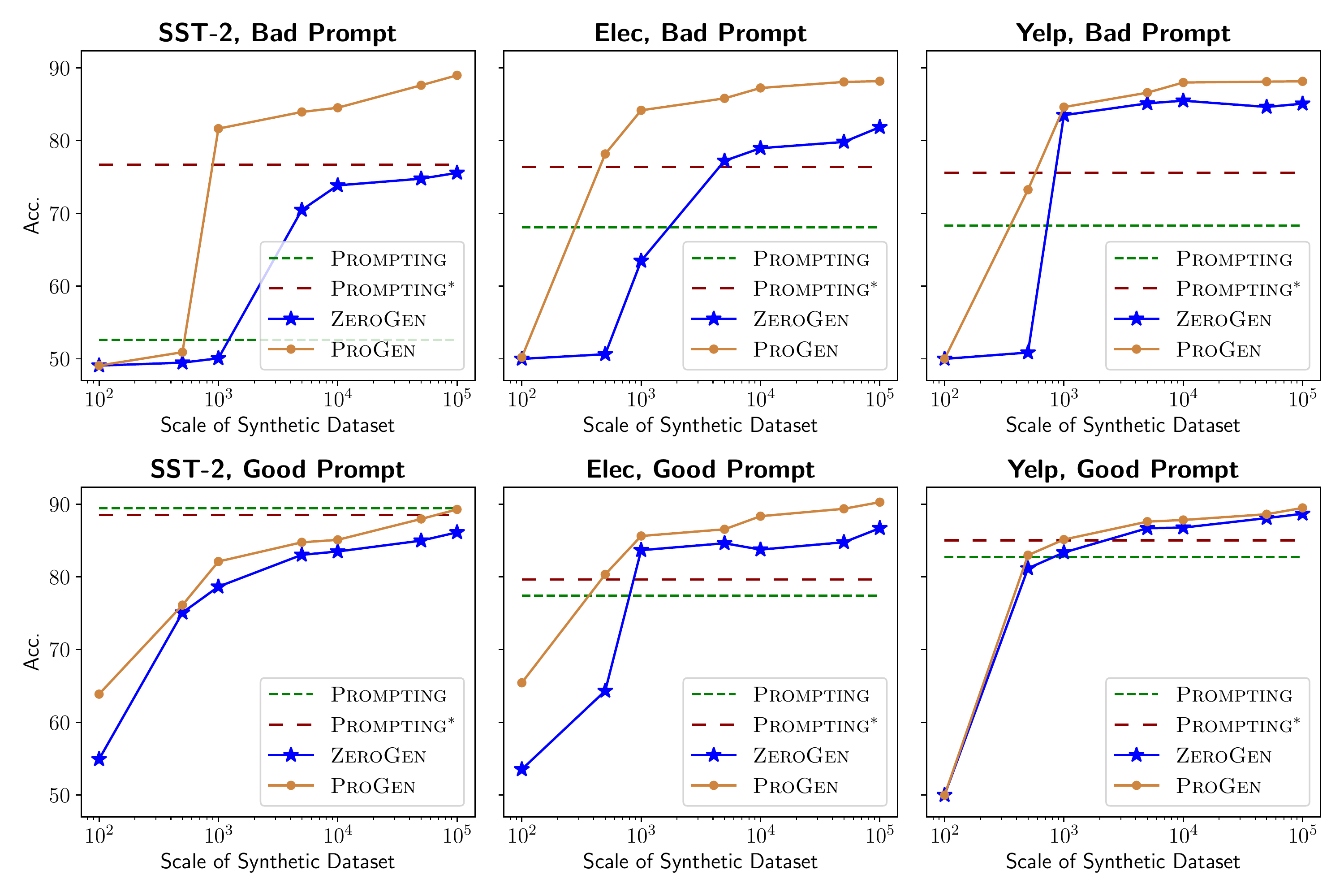}
\caption{Zero-shot performance with TAM trained under various scale of synthetic dataset.
}
\label{fig:scale}
\end{figure*}

Given the identified important examples, it's also unknown how to organize these examples as in-context examples. Previous work suggests the order of these examples plays a key role in model performance~\citep{DBLP:conf/acl/KumarT21,DBLP:conf/acl/LuBM0S22}, and the performance improves as the number of in-context examples increase~\citep{DBLP:conf/nips/BrownMRSKDNSSAA20}. However, these effects are still under-explored in zero-shot dataset generation. 

Suppose we have identified a bunch of positive and negative important examples, and are going to generate with a positive sentiment prompt (e.g., "\textit{The movie review in positive sentiment is: "}"), we study the following formats of in-context examples:
\begin{compactitem}
\item \textbf{Base}: no in-context examples (\textsc{ZeroGen}).
\item \textbf{F-1}: positive and negative examples are randomly placed. 
\item \textbf{F-2}: positive examples are placed {before} negative. 
\item \textbf{F-3}: positive examples are placed {after} negative. 
\item \textbf{F-4}: only positive examples are placed.  
\item \textbf{F-5}: only positive examples are placed, but the label information is not expressed (e.g., using prompt "\textit{The movie review is: "<X>"}" for each in-context example, where <X> is the text placeholder to fill in for each in-context example.).
\end{compactitem}
Besides, we also study the number of in-context examples, and the results are shown in Figure~\ref{fig:format}. We have the following observations. First, a modest number of in-context examples (e.g., no more than 8) consistently improves the performance, however, more in-context examples do not always turn into better performance. We leave the study on larger PLMs (e.g., GPT-3) and PLMs supporting more in-context examples as future work. Second, the performance of different orders of positive and negative examples varies in different tasks, while masking the label information (i.e., {F-5}) consistently improves performance. 

\paragraph{Prompt selection is less important for \textsc{ProGen} while scaling dataset is still valuable.}
In this part, we visualize the performance change over two representative prompts (i.e., a good prompt $P_2$ and a relatively bad prompt $P_1$) and various scales of dataset, and the comparison is shown in Figure~\ref{fig:scale}.

Overall, we find that \ourmodel is more effective for the bad prompt than the good one, and the bad prompt can achieve comparable results with the good one with \ourmodel on all the datasets. This indicates the quality of dataset can be iteratively improved with previous lower-quality dataset slice, which shares a similar spirit with Self-Training~\citep{lee2013pseudo} that also learns from its own predictions. Besides, we find \ourmodel can achieve similar or superior performance to \textsc{ZeroGen} with only 1\% (100k vs. 1k)
size of synthetic dataset. This becomes more meaningful when we only have restricted access to PLMs in real-world applications.

\paragraph{In-depth analysis of the synthetic dataset.}
In previous sections, we measure the quality of the synthetic dataset by training a TAM with that dataset and evaluating on downstream human-annotated data. In this section, we provide other measurements for a more comprehensive understanding of the synthetic dataset. We measure $\mathcal{D}$ from two perspectives, i.e., texts and labels. Regarding texts, we use Self-BELU~\citep{DBLP:conf/sigir/ZhuLZGZWY18} to measure its own diversity and MAUVE~\citep{DBLP:conf/nips/PillutlaSZTWCH21} to measure the similarity between prediction distribution and ground-truth distribution. We measure the correctness of labels with an  oracle model trained on human-annotated data as done in~\citet{DBLP:journals/corr/abs-2202-07922}. The comparison of different metrics on the Elec dataset is reported in Table~\ref{tab:other_metrics}.

Firstly, we find \textsc{ZeroGen} achieves the highest diversity, and \ourmodel degrades diversity. This indicates PLMs can generate texts similar to in-context examples, and the high Self-BLEU score for \ourmodel (Gold) is mainly due to the limited number of in-context examples (i.e., 38k for Elec test set vs. 100k for synthetic dataset). 
Secondly, \ourmodel well shifts the generation distribution towards ground-truth distribution, and \ourmodel (Gold) achieves significantly higher MAUVE scores. 
Finally, both \ourmodel and \ourmodel (Gold) increase label correctness, which is also highly reflected in TAM.
Overall, providing feedback improves synthetic datasets in both text distribution and label correctness, but also sightly decreases diversity.

\begin{table}[t]
\centering
\small
\scalebox{0.72}{
\begin{tabular}{l|cccc}
\toprule
 & \multicolumn{1}{l}{\textbf{Self-BLEU}} & \multicolumn{1}{c}{\textbf{MAUVE}} & \multicolumn{1}{c}{\textbf{Correctness}} & \multicolumn{1}{c}{\textbf{TAM}} \\
Granularity & $f(\mathcal{X})$ & $f(\mathcal{X},\hat{\mathcal{X}})$ & $f(\mathcal{X},\mathcal{Y},\hat{\mathcal{X}},\hat{\mathcal{Y}})$ & $f(\mathcal{X},\mathcal{Y},\hat{\mathcal{X}},\hat{\mathcal{Y}})$ \\
 \hline
\textit{Prompt} $P_1$ & & & & \\
\textsc{ZeroGen} & \textbf{19.46} & 79.13 & 62.44 & 81.85 \\
\ourmodel & 20.70 & 81.69 & 74.48 & 88.00 \\
\ourmodel (Gold) & 27.37 & \textbf{92.45} & \textbf{89.01} & \textbf{90.67} \\
\hline
\textit{Prompt} $P_2$ & & & & \\
\textsc{ZeroGen} & \textbf{15.98} & 68.99 & 79.64 & 86.62 \\
\ourmodel & 17.66 & 74.64 & 80.86 & 90.27 \\
\ourmodel (Gold) & 28.12 & \textbf{94.34} & \textbf{87.61} & \textbf{90.87} \\
\hline
\textit{Prompt} $P_3$ & & & & \\
\textsc{ZeroGen} & \textbf{16.08} & 77.61 & 82.99 & 87.58 \\
\ourmodel & 19.80 & 80.55 & 84.30 & 88.72 \\
\ourmodel (Gold) & 26.16 & \textbf{93.75} & \textbf{89.59} & \textbf{91.52} \\
\bottomrule
\end{tabular}}
\caption{Quality of various generated datasets measured by metrics in different granularity. \ourmodel (Gold) refers to selecting in-context examples from gold test set. $\mathcal{X}$ and $\hat{\mathcal{X}}$ represent synthetic set and test set, respectively.}
\label{tab:other_metrics}
\end{table}

\section{Related Work}
\subsection{Dataset Generation with PLMs}
The accuracy of neural models highly depends on the availability of large-scale human-annotated training data, which, however, can be prohibitively expensive to obtain at scale. Recent advances in generative language models~\citep{radfordlanguage,DBLP:conf/nips/BrownMRSKDNSSAA20} arouse great interests on generating synthetic dataset with PLMs. Some works generate data with a generative model fine-tuned on the public human-annotated dataset
~\citep{DBLP:conf/aaai/Anaby-TavorCGKK20,DBLP:conf/emnlp/PuriSSPC20,DBLP:journals/corr/abs-2003-02245,DBLP:journals/corr/abs-2102-01335}. Regarding the low quality generations, sample selection~\citep{DBLP:conf/emnlp/YangMFSBWBCD20,DBLP:journals/corr/abs-2201-05955} have also been used as postprocessing, which is complementary to our method that improves the dataset quality during generation.

In the context of zero-shot dataset generation, previous approaches adopt prompt-based methods~\citep{DBLP:journals/corr/abs-2012-00955,DBLP:conf/emnlp/ShinRLWS20,DBLP:journals/corr/abs-2109-07830} to generate data without any human-annotations~\citep{DBLP:conf/emnlp/SchickS21a,meng2022generating,DBLP:journals/corr/abs-2202-07922,DBLP:journals/corr/abs-2205-12679}. The synthetic dataset can be used to train a task-specific model and perform zero-shot inference on downstream tasks.
In contrast to our work, all the previous works generate the whole dataset at once, while we consider the quality of previously generated 
instances and improve the dataset quality during generation. 

\subsection{In-context Learning}
\citet{DBLP:conf/nips/BrownMRSKDNSSAA20} suggest that large PLMs can learn a task by conditioning on a few input-output demonstration pairs as prompt. This paradigm, known as \textit{In-context learning}, is especially attractive as it eliminates the need for updating parameters of the large language model. 
Subsequent works include better ways of choosing in-context examples~\citep{DBLP:conf/acl-deelio/LiuSZDCC22,DBLP:conf/acl/LuBM0S22,DBLP:journals/corr/abs-2112-08633}, learning with an in-context learning objective~\citep{DBLP:conf/acl/MinLHZ22,DBLP:conf/acl/ChenZZK022}, empirical analysis of why in-context learning works~\citep{DBLP:journals/corr/abs-2202-12837}, theoretical analysis that in-context learning can be formalized as Bayesian inference~\citep{DBLP:journals/corr/abs-2111-02080}, and explorations on other tasks (e.g., semantic parsing~\citep{DBLP:conf/emnlp/PasupatZG21}, dialogue state tracking~\citep{DBLP:journals/corr/abs-2203-08568,DBLP:journals/corr/abs-2201-05966}). To the best of our knowledge, all previous works study in-context learning in a few-shot learning setting. In contrast, this work focuses on a zero-shot learning setting for dataset generation task and the PLMs' ability to learn from in-context synthetic important examples to produce better dataset.


\section{Conclusions}
This work proposes \ourmodel for zero-shot dataset generation, which progressively improves the dataset quality by leveraging feedback from a task-specific model trained on the current dataset. By evaluating zero-shot performance with the trained model, we show \ourmodel can generate a much smaller (e.g., 1\%) synthetic high-quality dataset that achieves comparable or superior performance to baseline method. We also provide a variety of analyses, including formats of in-context examples, other measurements on synthetic datasets, and the influence of prompt selection. 

\section*{Limitations}
Our work depends on the PLMs' following abilities: (1) learning from in-context examples; 
(2) generating relatively high-quality data when using only the  manual prompt. 
This means that if the task can not be well described by a prompt or the PLM is not exposed to enough task-related data in the pre-training stage, the progressive dataset generation process may fail due to the extremely low-quality initial dataset slice and validation set. It can also affect the task-specific model’s ability to identify important examples: a noisy validation data point can fool the classifier into trusting mislabeled examples that fall close to it, further degrading the generation quality.
In addition, calculating influence function in practice suffers from low-efficiency issues. In practice, we sample a subset from the entire synthesized dataset in each iteration to reduce the computation, which can be sub-optimal in quality estimation accuracy. 

\section*{Acknowledgement}
We thank the anonymous reviewers whose suggestions helped clarify this work. This work is partially supported by the Shanghai Committee of Science and Technology (Grant No.~21DZ1100100), and the joint research scheme of the National Natural Science Foundation of China (NSFC) and the Research Grants Council (RGC) under grant number N\_HKU714/21.

\bibliography{anthology}
\bibliographystyle{acl_natbib}

\appendix

\section{Prompt Design}
\label{app:prompt}

\begin{table*}[t]
\centering
\renewcommand\arraystretch{1.2}  
\scalebox{0.8}{
\begin{tabular}{l|lll}
\toprule
\textbf{Setting} & \textbf{Id} & \textbf{Prompt Details} & \textbf{Label word <Y>} \\
\hline
\multirow{2}{*}{\textsc{Prompting}\,/\,\textsc{Prompting}$^*$} & $P_1$ & <Y> <TASK> Review: "<X>" & Positive/Negative \\
 & $P_2$ & The <TASK> review in <Y> sentiment is: "<X>" & positive/negative \\
 \hline
{\multirow{3}{*}{\textsc{ZeroGen}\,/\,\ourmodel}} & $P_1$ & <Y> <TASK> Review: " & Positive/Negative \\
& $P_2$ & The <TASK> review in <Y> sentiment is: " & positive/negative \\
& $P_3$ & The <TASK> review in <Y> sentiment for <TASK> "<C>" is: " & positive/negative \\
\bottomrule
\end{tabular}}
\caption{
Multiple text prompts for each setting. "<X>" refers to the test input for \textsc{Prompting} setting. "<TASK>" represents "movie", "electronic product" and "restaurant" for IMDb/SST-2/Rotten Tomatoes, Elec, Yelp datasets, respectively. "<C>" is a generated text formulated as additional condition  to steer generation. 
}
\label{tab:prompt}
\end{table*}

Table~\ref{tab:prompt} summarizes the text prompts used in this work. We choose prompts with the most representative "Control Code" and "Natural Language" styles as discovered by~\citet{DBLP:journals/corr/abs-2202-07922}. We also include a two-stage prompt $P_3$ that uses an additional generated text by PLM as condition, e.g., the movie name is generated by prompt "Movie: "".

\section{Detailed Results on Each Prompt}
\label{app:full_results}

\begin{table*}[t]
\centering
\scalebox{0.9}{
\begin{tabular}{ll|cccccc}
\toprule
\textbf{Setting} & \textbf{Prompt} & \textbf{IMDB} & \textbf{SST-2} & \textbf{Rotten Tomato} & \multicolumn{1}{l}{\textbf{Elec}} & \multicolumn{1}{l}{\textbf{Yelp}} & \multicolumn{1}{l}{\textbf{Avg.}} \\
\hline
\textsc{Supervised} & - & 87.24 & 89.68 & 83.67 & 92.63 & 95.42 & 89.73 \\
\hline
\multirow{2}{*}{\textsc{Prompting}} & $P_1$ & 60.36 & 52.64 & 52.91 & 68.08 & 68.33 & 60.46 \\
 & $P_2$ & 80.64 & 89.45 & 84.24 & 77.44 & 82.70 & 82.89 \\
 \hline
 \multirow{2}{*}{\textsc{Prompting}$^*$} & $P_1$ & 
 75.73	& 76.72	& 73.55	& 76.41	& 75.57	& 75.60 \\
 & $P_2$ & 78.89 & 88.53 & 83.77 & 79.65 & 85.03 & 83.17 \\
 \hline
\multirow{3}{*}{\textsc{ZeroGen}} & $P_1$ & 74.20 & 75.57 & 69.50 & 81.85 & 85.08 & 77.24 \\
 & $P_2$ & 83.27 & 86.12 & 82.46 & 86.62 & 88.67 & 85.43 \\
 & $P_3$ & 83.76 & 86.61 & 83.11 & 87.58 & 89.76 & 86.16 \\
 \hline
\multirow{3}{*}{\ourmodel} & $P_1$ & 84.22 & 87.16 & 83.02 & 88.00 & 89.06 & 86.29 \\
 & $P_2$ & 84.31 & 86.01 & 81.52 & 90.27 & 89.48 & 86.32 \\
 & $P_3$ & 83.82 & 88.42 & 84.05 & 88.72 & 89.63 & 86.93 \\
\bottomrule
\end{tabular}}
\caption{Results on each prompt with DistilBERT as task-specific model.}
\label{tab:full_bert}
\end{table*}

\begin{table*}[t]
\centering
\scalebox{0.9}{
\begin{tabular}{ll|cccccc}
\toprule
\textbf{Setting} & \textbf{Prompt} & \textbf{IMDB} & \textbf{SST-2} & \textbf{Rotten Tomato} & \multicolumn{1}{l}{\textbf{Elec}} & \multicolumn{1}{l}{\textbf{Yelp}} & \multicolumn{1}{l}{\textbf{Avg.}} \\
\hline
\textsc{Supervised} & - & 84.60 & 76.30 & 77.49 & 86.36 & 91.30 & 83.21 \\
\hline
\multirow{2}{*}{\textsc{Prompting}} & $P_1$ & 60.36 & 52.64 & 52.91 & 68.08 & 68.33 & 60.46 \\
 & $P_2$ & 80.64 & 89.45 & 84.24 & 77.44 & 82.70 & 82.89 \\
 \hline
  \multirow{2}{*}{\textsc{Prompting}$^*$} & $P_1$ & 
 75.73	& 76.72	& 73.55	& 76.41	& 75.57	& 75.60 \\
 & $P_2$ & 78.89 & 88.53 & 83.77 & 79.65 & 85.03 & 83.17 \\
 \hline
\multirow{3}{*}{\textsc{ZeroGen}} & $P_1$ & 60.33 & 63.88 & 61.73 & 71.16 & 81.29 & 67.68 \\
 & $P_2$ & 74.80 & 80.50 & 76.92 & 74.12 & 84.58 & 78.18 \\
 & $P_3$ & 75.40 & 82.22 & 78.80 & 82.24 & 85.39 & 80.81 \\
 \hline
\multirow{3}{*}{\ourmodel} & $P_1$ & 77.43 & 81.08 & 76.64 & 79.21 & 84.44 & 79.76 \\
 & $P_2$ & 77.30 & 79.13 & 76.17 & 84.38 & 85.97 & 80.59 \\
 & $P_3$ & 78.81 & 82.68 & 78.99 & 84.96 & 87.68 & 82.62 \\
 \bottomrule
\end{tabular}}
\caption{Results on each prompt with LSTM as task-specific model.}
\label{tab:full_lstm}
\end{table*}
The detailed results on each prompt are reported in Table~\ref{tab:full_bert} and Table~\ref{tab:full_lstm}.

\section{Robust Influence Function on Artificial Noisy Data}
\label{app:artificial}

\begin{figure}[t]
\centering
\includegraphics[width=3in]{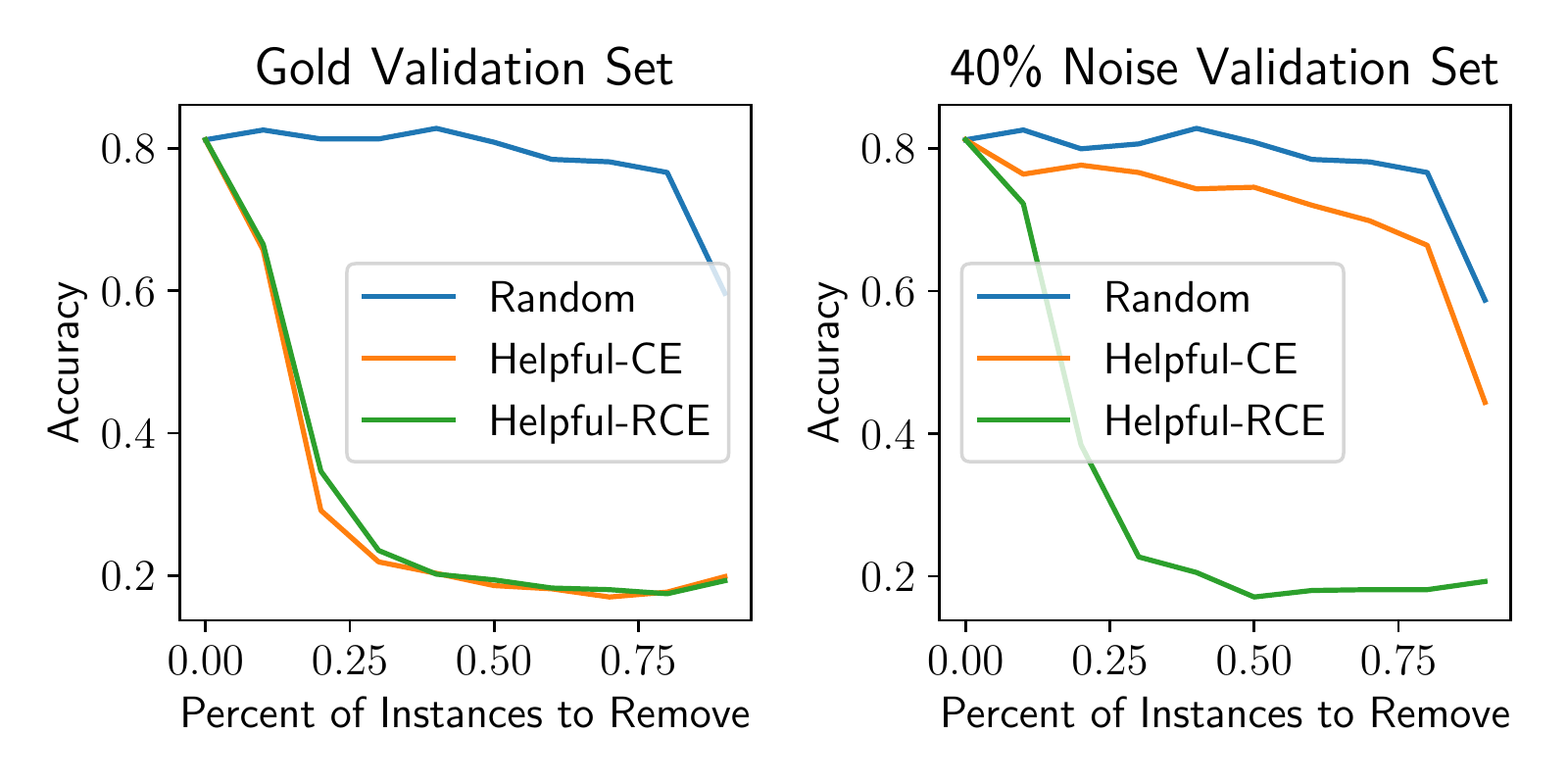}
\caption{Comparison on artificial noisy dataset. 
}
\label{fig:if-appendix}
\end{figure}

To investigate the ability of robust influence function in a noisy-validation set scenario, we create an artificial noisy dataset based on the human-annotated dataset. Specifically, we reverse a portion (e.g., 40\%) of ground-truth labels in the training and validation set, and compare the results of using gold validation set and mislabeled validation set when calculating influence function. We show the result comparison in Figure~\ref{fig:if-appendix}, where 
we remove examples based on the calculated score, retrain the model with the rest of dataset, and evaluate it on a held-out human-annotated set. 
We find when the validation set is well labeled, both cross entropy-based and reverse cross entropy-based influence function achieve on-par performance. However, a noisy validation set with cross-entropy has a great impact on the quality estimation -- removing identified helpful examples only slightly degrades accuracy. In contrast, a noise-resistant objective still provides reliable estimates.

\end{document}